# Cerberus: A Deep Learning Hybrid Model for Lithium-Ion Battery Aging Estimation and Prediction Based on Relaxation Voltage Curves

Yue XIANG,[1] Bo JIANG,[1*] Haifeng DAI [1*]

1 Clean Energy Automotive Engineering Center, School of Automotive Engineering, Tongji University, 201804 Shanghai, China.
* Corresponding author.

Yue Xiang: 2131536@tongji.edu.cn, Bo Jiang: jiangbo15@tongji.edu.cn[*], Haifeng Dai: tongjidai@tongji.edu.cn[*].

## Abstract

The degradation process of lithium-ion batteries is intricately linked to their entire lifecycle as power sources and energy storage devices, encompassing aspects such as performance delivery and cycling utilization. Consequently, the accurate and expedient estimation or prediction of the aging state of lithium-ion batteries has garnered extensive attention. Nonetheless, prevailing research predominantly concentrates on either aging estimation or prediction, neglecting the dynamic fusion of both facets. This paper proposes a hybrid model for capacity aging estimation and prediction based on deep learning, wherein salient features highly pertinent to aging are extracted from charge and discharge relaxation processes. By amalgamating historical capacity decay data, the model dynamically furnishes estimations of the present capacity and forecasts of future capacity for lithium-ion batteries. Our approach is validated against a novel dataset involving charge and discharge cycles at varying rates. Specifically, under a charging condition of 0.25C, a mean absolute percentage error (MAPE) of 0.29% is achieved. This outcome underscores the model's adeptness in harnessing relaxation processes commonly encountered in the real world and synergizing with historical capacity records within battery management systems (BMS), thereby affording estimations and prognostications of capacity decline with heightened precision.

## Introduction

Lithium-ion batteries exhibit significantly higher energy and power densities compared to traditional lead-acid and alkaline batteries, rendering them indispensable in applications ranging from electric vehicles to the 3C electronics domain.[1, 2] Despite the substantial increase in the cycle life of lithium-ion batteries over conventional counterparts, performance degradation remains inevitable due to a complex interplay of mechanical, electrical, and chemical factors, both internal and external.[3, 4] Hence, there exists a pervasive interest in accurately and promptly estimating or predicting the aging state of lithium-ion batteries. State of Health (SoH), quantified by the current maximum charge and discharge capacity, stands out as a prominent metric to gauge the level of degradation.[5] Swiftly estimating the SoH of in-service lithium-ion batteries and, furthermore, precisely forecasting the trajectory of their SoH degradation in the future, holds paramount

significance for optimizing the performance and utilization of lithium-ion batteries.[6]

Presently, methodologies for estimating and predicting SoH can be broadly categorized into two classes: model-based methods (primarily encompassing electrochemical models and equivalent circuit models) and data-driven approaches.[7, 8] Electrochemical methods accomplish SoH estimation and prediction through an analysis of the battery's degradation mechanisms from an electrochemical perspective, while equivalent circuit models predominantly employ control theory-based modeling of external battery characteristics to analyze aging.[9, 10] In contrast, data-driven approaches circumvent the need to account for the intricate electrochemical variations within lithium-ion batteries, relying on extensive data to extract parameters and features indicative of battery aging and subsequently fitting the nonlinear mapping between these features and SoH.[11] The data-driven models, owing to their independence from explicit knowledge of electrochemical parameters, ease of deployment, and robustness to variations in initial parameters, have emerged as a focal point in the current landscape of lithium-ion battery aging estimation and prediction.[12, 13]

Within the domain of data-driven lithium-ion battery state estimation, a diverse array of methodologies has surfaced, each extracting distinct aging features or adopting varying model structures. Yang et al. have derived four geometric features from CC-CV charging curves and employed Gaussian Process Regression (GPR) for SoH estimation, yielding an approximate RMSE of 6% on a set of randomized charge-discharge tests.[14] The limited precision of this approach is likely attributable to the inadequate generalization capability of the chosen features under intricate operating conditions. Lyu et al., on the other hand, have harnessed two aging features from partial differential capacity curves during the charging process, devising a Gaussian Process Particle Filter (GPPF) to estimate battery capacity with a maximum absolute percentage error within 2.91%.[15] Li et al. have introduced a semi-supervised Transfer Component Analysis approach, which leverages a KRR model for battery SoH estimation.[16] This model employs a method to eliminate redundant information and minimize disparities between distinct data distributions, facilitating improved estimation outcomes.

Predicting the state of lithium-ion batteries through data-driven approaches presents distinct challenges, often proving arduous to concurrently achieve high predictive accuracy and robust generalization capabilities. Zhang et al. have addressed this by employing two methods, namely correlation coefficients and decision trees, to sift initial features from raw charge-discharge data.[17] Subsequently, they undertake further refinement via the Variance Inflation Factor (VIF) and devise a residual capacity prediction framework within the Gradient Boosting Decision Tree (GBDT) model framework. Yang et al. employ the Hybrid Pulse Power Characterization (HPPC) test-based direct parameter extraction technique to identify battery parameters for a first-order equivalent circuit model.[18] These parameters are then utilized to train a three-layer Backpropagation Neural Network (BPNN) for SoH prediction, demonstrating computational efficiency in the process. Hong et al., alternatively, delve into an extensive exploration of deep learning models, considering terminal voltage, current, and battery temperature.[19] Their proposed framework incorporates uncertainty metrics, t-SNE feature analysis, and diverse battery-relevant tasks to comprehensively analyze the deep learning model. The resultant framework significantly enhances predictive accuracy in remaining cycle life estimation, yielding a noteworthy mean absolute percentage error (MAPE) of 10.6%.

Existing research predominantly fixates on a singular aspect of estimating or predicting the degradation state of lithium-ion batteries, failing to effectively unify the two endeavors. On one hand, due to the scarcity of opportunities for electric vehicles to undergo comprehensive SoH capacity calibration in routine operation, historical aging prediction methods struggle to acquire authentic and accurate training data and labels. Conversely, the outcomes of capacity estimation are confined to battery retirement or recycling alerts, without harnessing the full potential of the nonlinear mapping encapsulating electrochemical mechanisms that the estimation models glean from the interplay between aging features and capacity.

This study introduces an integrated model for SoH aging estimation and prediction of lithium-ion batteries, termed the Cerberus model. This innovative framework capitalizes on relaxation voltage curves from charge and discharge processes, in conjunction with historical capacity estimation outcomes. Employing a multi-layered Bidirectional Gated Recurrent Unit (bi-GRU), a unidirectional Long Short-Term Memory (LSTM), and a multi-layer perceptron as the foundational architecture, the Cerberus model employs a sliding window weighted averaging technique to reconcile distinct accuracy requirements during the early and later phases of aging. This strategy ensures a harmonious equilibrium between the precision demands of estimation and prediction while mitigating model complexity. Moreover, the model's requisite training data can be sourced from routine on-board Battery Management System (BMS) records, thus underscoring its high practical utility and marking a substantial stride towards the realization of a comprehensive lithium-ion battery degradation twin model.

## Methods

**Data source**

The lithium-ion battery aging data utilized in this study are sourced from the publicly available dataset titled *Data-driven capacity estimation of commercial lithium-ion batteries from voltage relaxation*.[13] The dataset encompasses a substantial collection of large-scale cycling data concerning NCA batteries. Specifically, 33 NCA batteries with a nominal capacity of 3500mAh are subjected to diverse charging current rates within a temperature-controlled chamber at 25°C. The charging and discharging currents span a range of 0.25 C to 4 C.

The relaxation process referred to herein denotes the phenomenon occurring after the battery's charge or discharge, during which the internal chemical reaction rate gradually adjusts to attain a new equilibrium state. Furthermore, the diffusion rate of lithium ions evolves since they necessitate time to redistribute themselves to adapt to the altered conditions. In this context, relaxation pertains to the post-charge or post-discharge zero-current quiescent period.

In a complete charging, discharging, and relaxation cycle, the NCA battery undergoes an initial constant current charging to an upper voltage limit of 4.2V, with current rates ranging from 0.25C to 1C. Subsequently, it is subjected to constant voltage charging at 4.2V until the current diminishes to 0.05C. Following this, a 30-minute quiescent period ensues as a charging relaxation phase. Subsequent to the relaxation period, the NCA battery is discharged at a constant current until reaching 2.65V, followed by another quiescent period to stabilize voltage, thus constituting the discharging relaxation phase. The capacity calculated during the constant current discharging process is considered as the SoH capacity of the battery for the cycling period. The sampling interval during the charging and discharging processes is set at 2 seconds, while the sampling interval for the relaxation process is 60 seconds. A detailed configuration of the dataset is presented in Table 1.

As depicted in Fig. 1C, the battery capacity exhibits conspicuous nonlinear decline with increasing cycles. Consistent with prior investigations, the charging rate significantly influences the lifespan of lithium-ion batteries. During the degradation process to 80% SoH, a pronounced linear characteristic is discerned in capacity decay under the 0.25C charging condition, whereas severe nonlinear capacity attenuation is evident for the 0.5C and 1C charging scenarios.

Fig. 1A portrays the evolution of charge relaxation voltage curves in response to SoH degradation. As the aging of lithium-ion batteries progresses, these curves gradually shift downward, indicative of the continual loss of active materials in both the positive and negative electrodes. In the advanced stages of aging, relaxation voltage curves may even exhibit relaxation voltage drops of up to 0.04V. In contrast, Fig. 1B illustrates the variation of discharge relaxation voltage curves with SoH decline. The stabilized voltage post-rebound substantially elevates with increasing cycling count, and the time taken to attain stability progressively lengthens. Notably, these alterations are more pronounced in the discharge relaxation process compared to the charge relaxation process.

In summary, both charge and discharge relaxation voltages exhibit a pronounced correlation with SoH. It has been posited that relaxation voltage curves closely relate to the increase in internal resistance of lithium-ion batteries, primarily stemming from losses in lithium inventory, degradation of active materials, and augmented internal resistance. Given the strong positive correlation between elevated internal resistance and SoH degradation, the selection of relaxation voltage curves as aging features in this study is well-founded.

**Table 1 | Dataset details**

| Cell type | Structure | Nominal capacity(mAh) | Cutoff voltage(V) | Charge/Discharge current rate (C) | Test temperature (°C) | Number of cells |
|---|---|---|---|---|---|---|
| NCA | 18650 | 3500 | 2.65~4.2 | 0.25/1 | 25 | 5 |
| | | | | 0.5/1 | | 19 |
| | | | | 1/1 | | 9 |

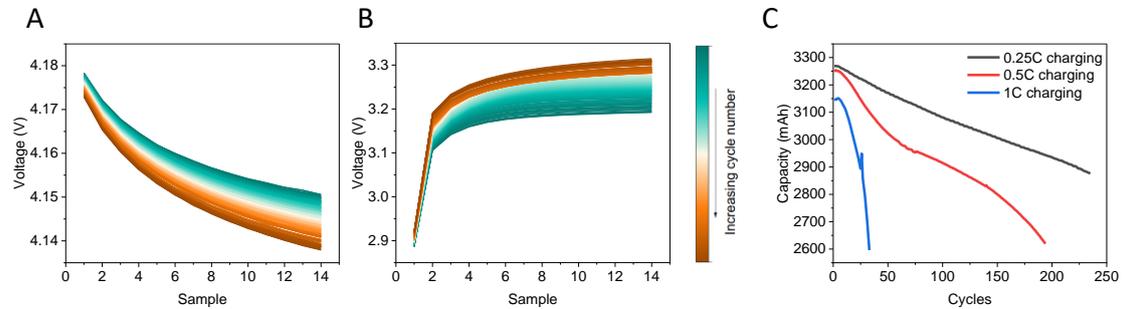

**Fig. 1 | Relaxation process and capacity degradation**

**Dataset split**

To enhance the model's robustness for adaptation to diverse operational conditions post-deployment, we employ a dataset comprising 33 NCA battery cells subjected to three distinct charging rate conditions at room temperature for both training and testing. In this study, two partitioning strategies are employed for the division of training and testing datasets: random partitioning and stratified partitioning.

a) Random Partitioning. For the random partitioning approach, relaxation voltage and capacity data across the three operational conditions are pooled, and an 80% subset is randomly designated as the training dataset, while the remaining 20% forms the testing dataset. Given that the relaxation voltage samples exhibit shorter intervals in the 0.25C charging configuration, and in consideration of real-world vehicle operation, downsampling to a 120-second interval is applied to these samples within the training set. This adjustment ensures uniformity in the length of relaxation process data across all conditions, which are then amalgamated as inputs for model training. To preempt potential issues with gradient explosion during the deep learning process, all input continuous voltage time sequences are normalized, thereby transforming them into z-scores. For evaluating estimation precision, the MAPE is computed on either the testing dataset or the entire combined training and testing dataset.

b) Stratified Sampling Partitioning. Due to the necessity of incorporating the battery's previous aging data for the aging prediction component, random partitioning is unsuitable. Consequently, in the stratified sampling approach, the historical aging capacity of the 33 lithium-ion batteries, representing the three charging scenarios, is employed for partitioning. In alignment with this, an 80% subset is designated as the training dataset, with the remaining 20% assigned to the testing dataset. During joint testing, the corresponding historical SoH windows and the concurrent relaxation voltage curves are synchronized for simultaneous estimation and prediction. The computation of the MAPE ensues as part of this evaluation process.

**Model Architecture**

As depicted in Fig. 2A, the comprehensive model comprises three principal components:

a) Extraction of aging features from post-charge relaxation phase: Following the acquisition of complete relaxation voltage curves, uniform downsampling to a 120s per interval is performed. Employing a sliding window technique, voltage time sequences are segmented into windows of 10 samples each (equivalent to 20 minutes per segment). These windows are incremented by 1 sample, and each window is associated with the discharge capacity label of the respective cycle. The charge time series data is then input into a two-layer bi-GRU. The resultant final hidden state output from this bi-GRU is subsequently connected to a three-layer Multilayer Perceptron (MLP) with layer sizes of 100-50-1. This configuration facilitates the learning of the nonlinear mapping between charge relaxation time sequences and SoH, concurrently serving to compress dimensions.

b) Extraction of aging features from post-discharge relaxation phase: Analogous to the procedures applied during the charge relaxation process, a similar methodology is employed here. However, due to the higher discharge current rates in the discharge process, a lengthier relaxation phase is required as an aging feature. Consequently, the window size is set to encompass 15 samples each (equivalent to 30 minutes per segment). Similarly, the discharge time series data is input into another two-layer bi-GRU. The ultimate hidden state output from this bi-GRU is then linked to a three-layer MLP with layer sizes of 100-50-1.

c) Prediction of future degradation trajectories from historical aging data: Upon acquiring discharge capacity curves for all 33 batteries, an expanded window approach with a fixed left endpoint is utilized to obtain the complete historical capacity degradation data for each battery. Linear extrapolation is employed for stages with insufficient or poor-quality early-stage aging data, ensuring a minimum of 10 battery cycles. This variable-length time series is input into a two-layer unidirectional LSTM. The ultimate cell state output from this LSTM is then connected to a three-layer MLP with layer sizes of 50-20-1.

Fig. 2B provides an in-depth depiction of the dual-layer bi-GRU structure. Both the GRU and LSTM are categorized as recurrent neural networks with gate mechanisms.[20] These gate mechanisms are designed to accumulate and forget historical information over extended periods, thereby addressing the issues of vanishing and exploding gradients during the training process. Notably, the bi-GRU considers both forward and reverse directions of a time series for analysis, thus encompassing the contextual relationships within aging voltage time sequences. Conversely, the unidirectional LSTM processes the lithium-ion battery capacity degradation information following the natural chronological order of aging. Prior research indicates a fundamental distinction in the fitting capability for nonlinear mappings between dual-layer recurrent neural networks and single-layer recurrent neural networks, with marginal variations arising from additional layers.[21] Consequently, we employ dual-layer Bidirectional Gated Recurrent Units in the 'a' and 'b' segments of the model, while a unidirectional LSTM is adopted for the 'c' segment. This choice strikes a balance between model capacity and training complexity. The entire network is trained on an NVIDIA GeForce GTX 1060 GPU.

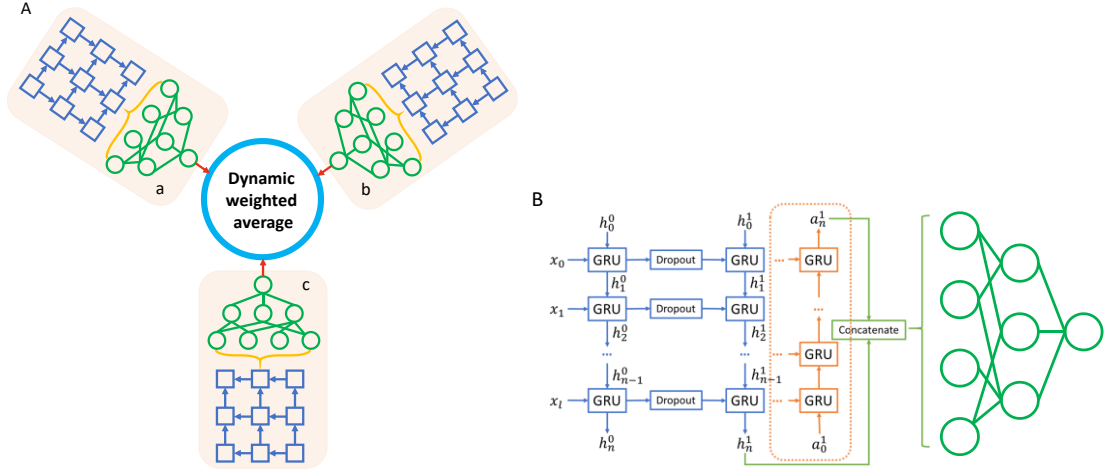

**Fig. 2 | Model structure**

**Formulation of the Loss Function**

A novel formulation of the loss function is introduced in this study to address the amalgamation of aging estimation and testing for lithium-ion batteries. Given the scarcity of historical data pertaining to degradation capacity in the early stages of battery degradation, a greater reliance on aging state information inferred from current relaxation voltage curves becomes imperative. Conversely, as an abundance of aging data becomes available in the advanced stages of degradation, the predictive component of model 'c' can offer more nuanced insights into the battery's distinctive characteristics. Consequently, a sliding window dynamic weighting approach is adopted, whereby distinct confidence factors—designated as α, β, and 1-α-β—are dynamically assigned to the three segments of results. As regression problems underpin all three segments, the Mean Squared Error (MSE), a well-established metric within the regression domain, is employed as the loss function. The comprehensive loss function is thus defined as follows:

$$\begin{aligned}\mathcal{L}_{total} &= \mathcal{L}_a + \mathcal{L}_b + \mathcal{L}_c \\ &= \alpha MSE(\bar{Y}_1, Y_1) + \beta MSE(\bar{Y}_2, Y_2) + (1-\alpha-\beta)MSE(\bar{Y}_3, Y_3)\end{aligned}$$

## Results

Fig. 3 illustrates the fused estimation and prediction outcomes of the Cerberus model proposed in this study. Fig. 3A depicts the comparison between the degradation, estimation, and prediction fusion values of one lithium-ion battery operating under 0.25C charge conditions against the actual values. Given the relatively regular linear aging characteristics observed in this scenario, the

Cerberus model exhibits an average estimation and prediction MAPE of approximately 0.29%. Fig. 3B reveals the model's performance under 0.5C charge conditions, characterized by an initial acceleration followed by deceleration in non-linear aging patterns. The model adeptly tracks these changing patterns, yielding an average estimation and prediction MAPE of around 4.38%. Despite a 5.14% MAPE error for the swift capacity drop observed in 1C charge conditions (Fig. 3C), the model remains capable of accurately following the degradation trend.

Fig. 3D provides a comparison between the Cerberus fused capacity estimation and prediction model proposed in this study against a solely relaxation process-based SoH estimation model or a historical capacity degradation-based prediction model. Across the three distinct charge conditions, the accuracy of the proposed model surpasses that of non-fused models. Particularly noteworthy is the substantial reduction in prediction MAPE compared to non-fused standard estimation models (3.91%, 3.57%) and prediction models (4.09%) when applied to the low-rate 0.25C charge condition. This underscores the efficacy of the dynamic weighted average method, adeptly balancing the limitations associated with inadequate prediction precision in the early stages of LIB degradation due to sparse historical aging data and the fluctuations observed in relaxation voltage curves during the advanced stages of aging. Even in the challenging 1C high-current charge scenario, the proposed model's MAPE is approximately 2% to 3% lower than the three baseline models. Notably, the estimation accuracy of the purely discharge relaxation process-based capacity estimation model outperforms that of the charge relaxation process-based model, potentially attributable to the more pronounced disparities observed in discharge relaxation voltage as SoH degrades.

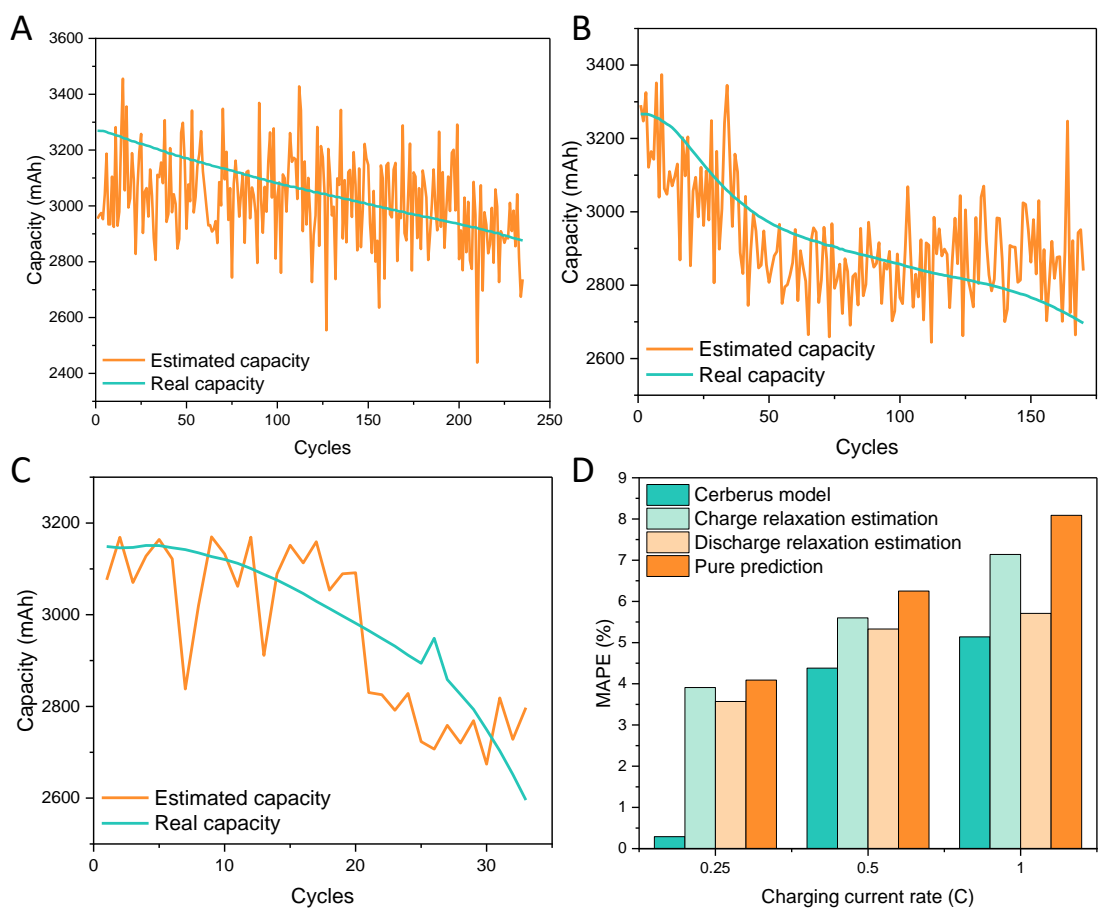

**Fig. 3 | Estimation and prediction results**

# Conclusion

As a crucial avenue for reducing greenhouse gas emissions and advancing clean energy, lithium-ion batteries have increasingly assumed pivotal roles in electric vehicles, energy storage systems, and consumer electronics. Consequently, the assessment perspectives and methodologies concerning the aging processes of lithium-ion batteries have become increasingly diverse. In this study, we introduce the Cerberus model, which employs a deep GRU architecture to extract aging-relevant mappings from relaxation voltage time series. Integrating this with aging trends derived from historical capacity data through a sliding window dynamic weighting approach, the Cerberus model unifies the estimation of the current aging state and the prediction of future degradation trajectories in lithium-ion batteries. The Cerberus model is trained and tested on a dataset encompassing 33 NCA-based battery systems. In conditions closely resembling real-world on-road scenarios, particularly at a 0.25C discharge rate mirroring typical single-cell charge rates, the model achieves a MAPE of under 0.3%. This performance underscores the model's capacity to balance estimation accuracy and robustness. Moreover, the model adeptly capitalizes on the commonplace relaxation processes that occur post-daily charging or battery usage, signifying its wide applicability in real-world contexts. As the demand for sustainable energy solutions grows, the Cerberus model contributes to the evolving landscape of lithium-ion battery aging assessment by providing a comprehensive and integrated approach.